\documentclass[journal]{IEEEtran}
\usepackage{blindtext}
\usepackage{graphicx}
\usepackage{draftwatermark}
\usepackage{epstopdf}
\usepackage{amsmath}
\usepackage{algorithm}
\usepackage{algorithmic}
\usepackage[export]{adjustbox}
\usepackage{tikz}
\usepackage{calc}
\usetikzlibrary{decorations.shapes}

\ifCLASSINFOpdf
\else
\fi

\SetWatermarkText{PREPRINT}
\SetWatermarkScale{1}
\begin{document}
\title{Image segmentation of cross-country scenes captured in IR spectrum}

\author{\IEEEauthorblockN{Artem Lenskiy}\\
\IEEEauthorblockA{Korea University of Technology and Education\\
1600, Chungjeol-ro, Byeongcheon-myeon,
Dongnam-gu, Cheonan-si, Chungcheongnam-do 31253,\\ Republic of Korea\\
email: a.a.lensky@gmail.com}
}

\maketitle

\begin{abstract}
Computer vision has become a major source of information for autonomous navigation of robots of various types, self-driving cars, military robots and mars/lunar rovers are some examples. Nevertheless, the majority of methods  focus on analysing images captured in  visible spectrum.  In this manuscript we elaborate on the problem of segmenting cross-country scenes captured in IR spectrum. For this purpose we proposed employing salient features. Salient features are robust to variations in scale, brightness and view angle. We suggest the Speeded-Up Robust Features as a basis for our salient features for a number of reasons discussed in the paper. We also provide a comparison of two SURF implementations. The SURF features are extracted from images of different terrain types. For every feature we estimate a terrain class membership function. The membership values are obtained by means of either the multi-layer perceptron or nearest neighbors. The features’ class membership values and their spatial positions are then applied to estimate class membership values for all pixels in the image. To decrease the effect of segmentation blinking that is caused by rapid switching between different terrain types and to speed up segmentation, we are tracking camera position and predict features' positions. The comparison of the multi-layer perception and the nearest neighbor classifiers is presented in the paper. The error rate of terrain segmentation using the nearest neighbors obtained on the testing set is 16.6$\pm$9.17\%.

\end{abstract}

\begin{IEEEkeywords}
autonomous ground vehicle, visual navigation, texture segmentation.
\end{IEEEkeywords}

\IEEEpeerreviewmaketitle

\section{Introduction}
Autonomouse navigation systems designed for autonomous driving in outdoor, non-urban environment is more complicated than a system designed for traversing in urban environment with plenty straight lines. A variety of perception systems have been proposed for aiding off-road navigation. Such systems process data obtained from varies sensors including laser-range finders, color and grayscale cameras. Each type of sensors provides its own advantages and disadvantages.

In this chapter we elaborate on the problem of segmenting cross-country scene images using texture information. Our approach takes an advantage of salient features that are robust to variations in scale, brightness and view angle. The discussed texture segmentation algorithm can be easily incorporated into navigation systems with laser range scanners and other types of sensors. Moreover, using salient features the computer vision system can be easily extended to object recognition simply by adding new types of objects into the training set.

The rest of the chapter is organized as follows. Section 2 overviews current perception algorithms employed in the field of autonomous navigation of unnamed ground vehicles in cross-country environment.  At the end of section 2, a flowchart of the proposed segmentation system is given. Section 3 compares two popular implementations of the speeded up robust features (SURF) algorithms. Section 4 describes feature detection part of the terrain segmentation. The proposed texture model is described in section 5 and the segmentation procedure is proposed in section 6. In section 7 we elaborate on 3D reconstruction for SURF features tracking. Experimental results and conclusions are given in sections 8 and 9 respectively.

\section{Literature overview}
The majority of systems for unmanned ground vehicles generally relay on the two types of sensors: cameras and laser-range scanners. The data obtained by the laser-range scanners is applied for precise 3D reconstruction of surrounding environment. From the reconstructed 3D scene it is possible to distinguish flat regions, tree trunks and tree crowns ~\cite{Lalonde2006}. Moreover, lase-range scanners are capable of operating at any time of the day. On the other hand one disadvantage of laser scanners is an inability to distinguish some types of terrain. For instance it is impossible to distinguish gravel, mud and asphalt as these terrains have similar point-cloud distributions. It is also hard to recognize either a reconstructed object is a tall patch of grass, a rock or a low shrub. Another possible disadvantage of 3D laser-range scanners is still a high price.

One of the sources of information that is commonly used for terrain segmentation and road detection is color. Color carries information that simplify discrimination of gravel and mud, or grass and rocks.  The recognition accuracy in this case depends on a surrounding illumination. To minimize the influence of variations in illumination, a representative training data set should be collected, that covers all expected environmental conditions. Additionally, the classifier should be able to adequately represent variabilities of perceived colors within each single class.  Manduchi et al. ~\cite{ Manduchi2006, Manduchi2005} proposed a classifier that estimates color density functions of each class by employing Gaussian Mixture Models. The motivation of this method  comes from the fact that, for the same terrain type, color distributions under direct and diffuse light often correspond to different Gaussians modes. To improve color based terrain type classification Jansen et al. ~\cite{Jansen2005} proposed a Greedy expectation maximization algorithm. The authors firstly clustered training images into environmental states, then before an input image is segmented its environmental state is determined. Color distributions vary from one environmental state to another even within the same terrain class. Overall, the authors were able to classify sky, foliage, grass, sand and gravel with the lowest probability of correct classification of 80\%. Nevertheless, such classes as grass, trees and bushes are still indistinguishable have similar colors, moreover color information is not available at night.

Texture is another characteristic employed for terrain segmentation. Texture features extracted from grayscale or color images makes it easy to separate grass, trees, mud, gravel, and asphalt. A number of computer vision algorithms have been successfully applied for terrain segmentation using various texture features. To classify texture features into six terrain classes, Sung et al. \cite{Sung2010} applied a two-layer perceptron with 18 and 12 neurons in first and second hidden layers respectively. The feature vector was composed of the mean and energy values computed for selected sub-bands of two-level Daubechies wavelet transform, resulting in 8 dimensional vector. These values were calculated for each of the three color channels. Thus, each texture feature contained 24 components. The experiments were conducted in the following manner. Firstly, 100 random  video frames were selected for extracting training patches. Then, among them ten were selected for testing purposes. The wavelet mean and energy were calculated for fixed 16x16 pixel sub-blocks. Considering a resolution of input images of 720x480 pixels, the sub-block of 16x16 pixels is too small to capture texture characteristics, especially at larger scales, which therefore leads to a poor texture scale invariance. They achieved the average segmentation rate of 75.1\%. Considering that color information was not explicitly used and only texture features were taken into account, the segmentation rate is promising, although there is still room for improvement.

Castano et al. ~\cite{Castano2001} experimented with two types of texture features. The first type of features is based on the Gabor transform, particularly the authors applied the Gabor transform with 3 scales and 4 orientations as described in~\cite{Lee1996}. The second type is relays on the histogram approach, where amplitudes of the complex Gabor transform is partitioned into windows and histograms are calcualted from the Gabor features for each window. The classifier for the first type of features modeled the probability distribution function using a mixture of Gaussians and performed a Maximum Likelihood classification. The second classifier represents local statistics by marginal histograms. Comparable performances were reached with the both models. Particularly, in the case when half of the hand segmented images were used for training and the other half for testing, the classification performance on the cross-country scene images was 70\% for mixtures of Gaussian and 66\% for histogram based classification. Visual analysis of the presented segmentation results suggests that the wrong classification happens due to a short range of scale independence of Gabor features, and rotational invariance that make texture less distinguishable.

Scenes captures in IR appear quite different compare to images captures in visible spectrum. The terrain appearance in IR spectrum is less affected by shadows and illumination changes. Furthermore, the reflection in visible light range  ($0.4-0.7\mu$m)  for grass and trees (birch, pine and fir) is indistinguishable. On the other hand there is a substantial difference in reflectance in infrared range.  By segmenting the surrounding environment outdoors, the robot is able to avoid trees and adjust the speed for traversal depending of the terrain types.
Kang et al. ~\cite{Kang2009} proposed a new type of texture features extracted from multiband images including color and near-infrared bands. Prior texture segmentation they reconstructed 3D scene using structure from motion module. The reconstructed scene was divided into four depth levels.  Depending on the level the appropriate mask size is selected. Overall they used 33 masks for each band constituting of 132 dimensional feature vectors. For each mask a product is calculated by multiplying pixels in the mask together according to their patterns. The obtained features are coined as higher-order local autocorrelation (HLAC) features ~\cite{Kurita1993, Otsu1988}. The authors presented segmentation results for four urban scenes, with the best segmentation recognition rates of 84.4\%, 76.1\%, 84.5\% and 76.1\%.  We tend to believe that the training set even with millions of features, i.e. a thousand features per image and thousands of images in the training set, will be very sparsely distributed in such a high dimensional feature space (132 dimensions). Therefore, classifiers will have hard time learning and generalizing from comparably small training sets in such high dimensional spaces.

The best segmentation accuracy is achieved then various types of sensor data and uncorrelated features are combined for segmentation. An interesting work has been performed by Blas et al. ~\cite{Blas2008,Konolige2009}. They combined color and texture descriptors for online, unsupervised cross-country road segmentation. The 27-dimensional descriptors were clustered to define 16 textons. Then each pixel is classified as belonging to one of them. During the segmentation process, a histogram for a 32x32 window is estimated. The histogram represents a number of occurred textons. The obtained histograms are clustered into 8 clusters. A histogram from each window is then assigned to one of the clusters. Unfortunately, the authors did not present quantitative segmentation results for cross-country terrain segmentation due to probably unsupervised nature of the algorithm. Lenskiy et al.~\cite{Lensky2008} also applied texton based approach for terrain recognition, however instead of features extracted from fixed size windows, they employed the SIFT features~\cite{Lowe1999}. The SIFT features relay on scale-space extrema detection for automatically detecting scales and location of interest points.

Rasmussen ~\cite{Rasmussen2002} provided a comparison of color, texture, distance features measured by the laser range scanners, and a combination of them for the purpose of cross-country scene segmentation. The segmentation was the worst when texture features were used alone. In the case when 25\% of the whole feature set was used in training, only 52.3\% of the whole feature set was correctly classified. One explanation of this poor segmentation quality is in the feature extraction approach. The feature vector consisted of 48 values representing responses of the Gabor filter bank. Specifically, it consisted of 2 phases with 3 wavelengths and 8 equally-spaced orientations. The 48-dimensional vector appears to have enough dimensions to accommodate a wide variety of textures, however as we mentioned above it is still high, considering that training set consisted of 17120 features. Besides the feature dimensionality, the size of texture patches also influenced the segmentation quality. The size of the patch was set a block of 15x15 pixels that is relatively small, which led to a poor scale invariance. Furthermore, features’ locations were calculated on the grid without considering an image content. Another reason of the problematic segmentation accuracy is in the low classifier’s capacity. The author chose a neural network with only one hidden layer with 20 neurons as a classifier. One layer feed-forward neural network is usually taking more iterations for partitioning concave clusters and often end up in a local minima. Considering terrain texture features are very irregular, neural networks with a higher number of layers is appropriate.

\begin{figure}[htp]
\centering
\includegraphics[trim=0.5cm 0.5cm 0.5cm 0.5cm,width=10cm,left]{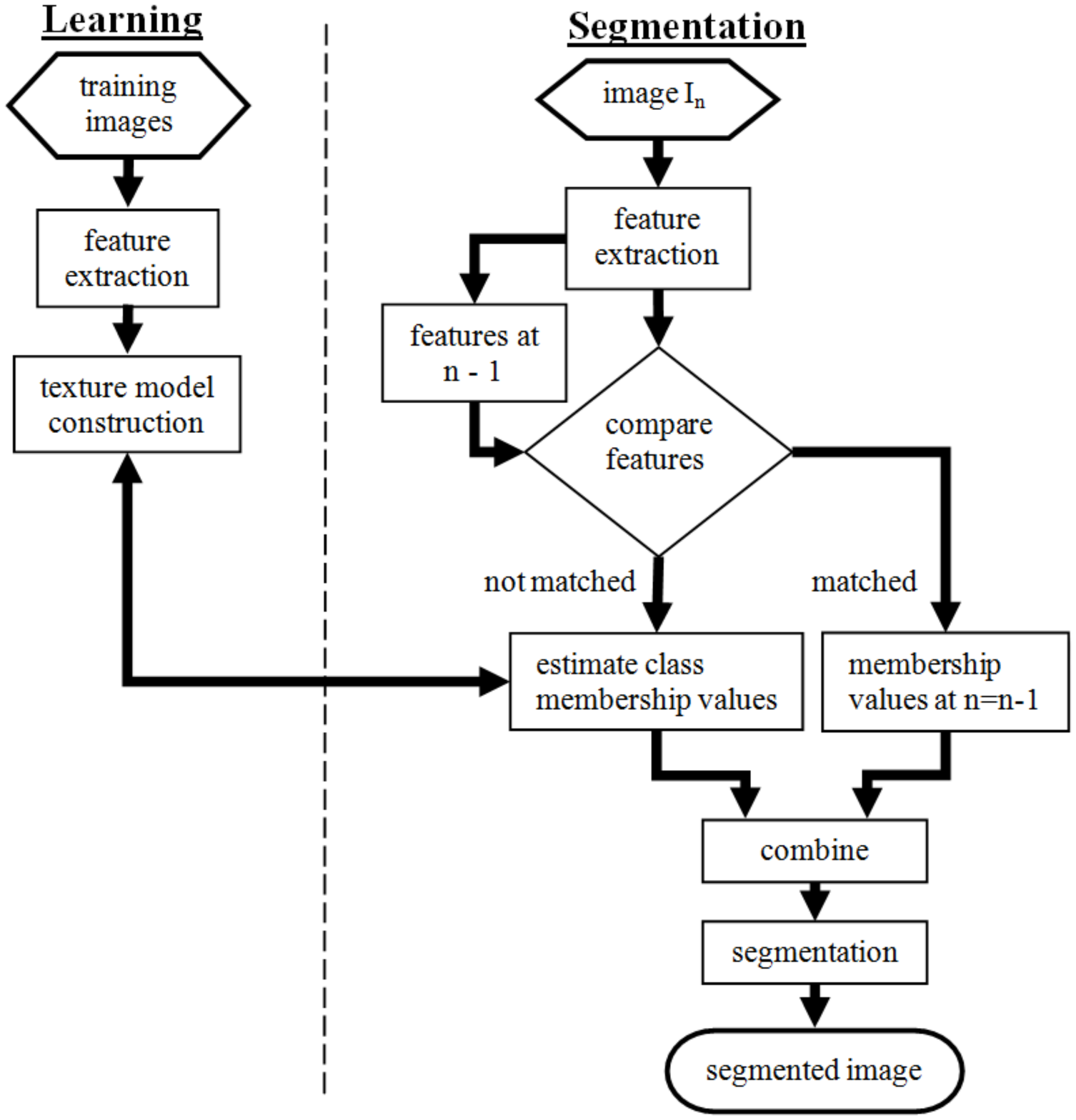}
\caption{The segmentation system block scheme}
\label{fig: scheme}
\end{figure}

In this chapter we elaborate on an early proposed texture segmentation algorithm that takes advantages of salient features ~\cite{Lenskiy2010a, Lenskiy2010b}. We present an analysis of the speeded up robust features \cite{Bay2008} for the purpose of terrain texture segmentation in IR images. For the reasons described above instead of 64 dimensional SURF we slightly change it implementation and reduce the dimensionality to 36 ~\cite{Lenskiy2011}. Salient features are detected in two steps: (1) features are localized and (2) descriptors are constructed. Features are searched at different image scales, i.e. size of image patches is not fixed that allows overcoming the problem mentioned above with fixed windows. The extracted features are applied in constructing texture models, that we apply for segment images (fig.~\ref{fig: scheme}). These features are extracted from a hand segmented images. After the texture model is constructed the system processes input video frames and returns segmented images, which is further used for navigating autonomous ground vehicle.

\label{Intro}

%\section{Learning stage}
\section{Comparison of SURF implementations}

The real-world texture segmentation are associated with two issues. The first one is a high variety of transformations affecting texture appearance. Even the same patch of texture, significantly changes it appearance under projective transformations including scale variations. Another issue is a high variation of textures within a class. For instance, the class corresponding to trees contains different textures for firs, pines etc.

One solution to the first problem is to find texture features invariant under projective transformation and brightness changes. A number of salient features have been proposed to satisfy such conditions. However, they have been applied mostly for object recognition and 3D reconstruction. Lenskiy and Lee  experimented with SURF features with different parameters and found that 36 dimensional SURF~\cite{Bay2008} features with no-rotational invariance (U-SURF) achieve the best segmentation accuracy on day time images~\cite{Lenskiy2011}. In this chapter we compare two SURF implementations and choose the one that suits our goals best.
\label{learning}

\begin{figure}
\centering
\includegraphics[width=\linewidth]{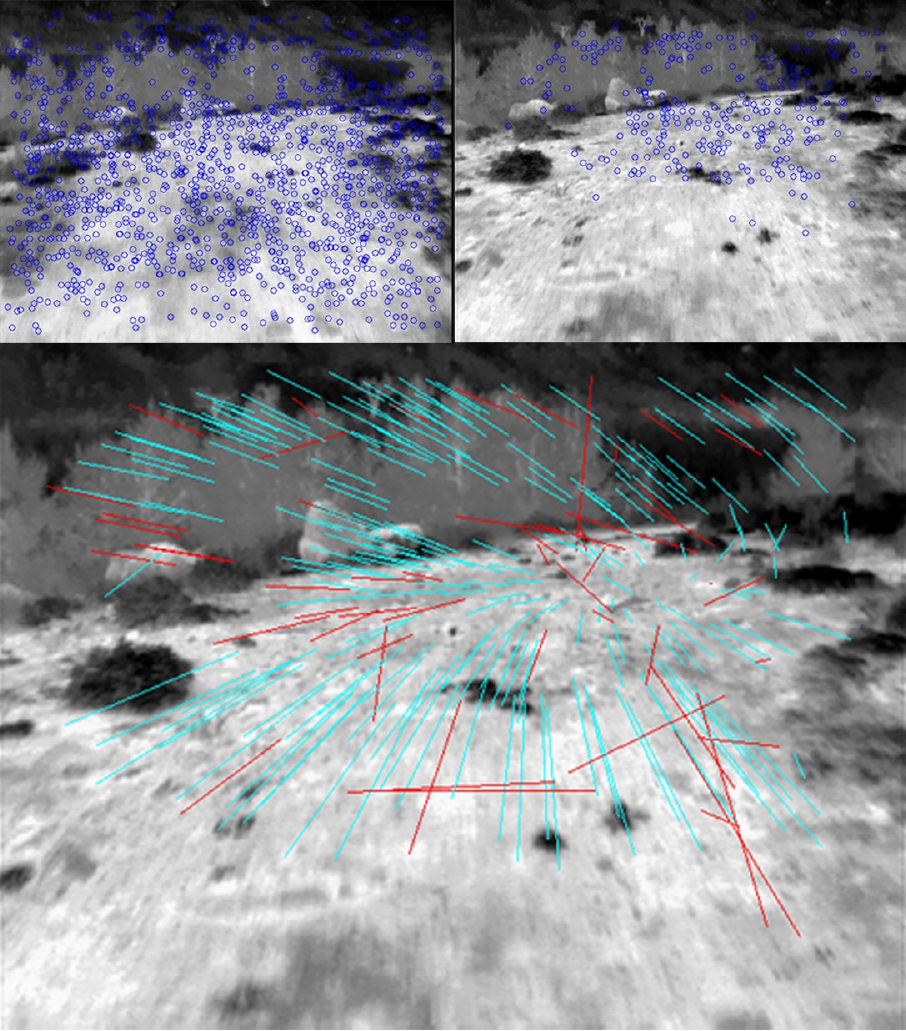}
\caption{Top left image shows detected SURF features, top right shows corresponding matched features for the second image, and in the bottom image cyan lines connects corresponding points, where red lines are outliers. }
\label{fig: detected_features}
\end{figure}

The \index{SURF algorithm}  SURF algorithm consists of three stages. In the first stage, interest points and their scales are localized in scale-space by finding the maxima of the determinant of the Fast-Hessian matrix. In the second stage, orientation is determined for each of the features prior to computing the feature descriptor. We use rotation dependent SURF, thus we omit this step. Finally a descriptor based on sampled Haar responses is extracted.

\begin{figure}
\centering
\includegraphics[width=\linewidth]{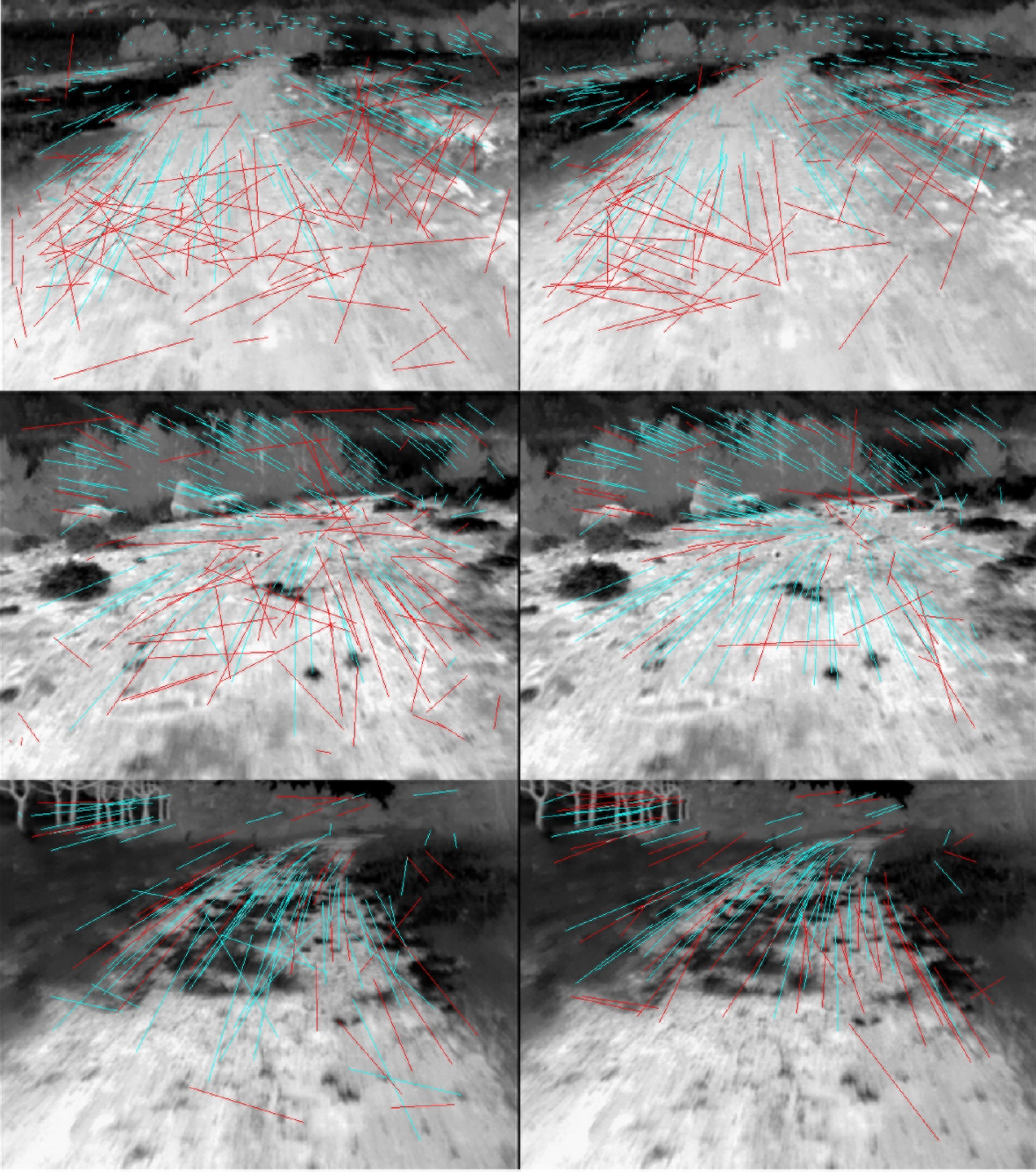}
\caption{Detected SURF features and their correspondences shown with cyan lines. Images in the left column are processed with OpenSURF and in the right with the original SURF implementations.}
\label{fig: correspondence features}
\end{figure}

The SURF algorithm has become a standard for salient features extraction. The SURF algorithm has been implemented for various hardware/software platforms such as Android, iPhone, Linux and others. A comparison of some open source implementations is given by Gossow et. al~\cite{Gossow2010}. The authors found that some implementations produce up to 33\% lower repeatability and up to 44\% lower maximum recall than the original implementation. We compare an open source implementation of the SURF named OpenSURF~\cite{Gossow2010} and the original SURF~\cite{Bay2008} implementations \index{SURF implementation comparison}. For this purpose we took five pairs images from a recorded video. Images in each pair are taken at different but close times, showing almost the same scene. Considering a slow vehicle movement, the images do not change much. We extract SURF features from each image and then matched them. Figure ~\ref{fig: correspondence features} shows three pair of images. The parameters for both SURF implementations are set identical. The comparison results are shown in table~\ref{tab: comparison}.  First column shows what implementation of the SURF algorithm is used to extract features. Second column in the table shows the number of detected features in each image in the analyzed pairs. Third column show the number of matched features. As it can be seen the number of matched features is significantly less than the number of detected features. By changing the parameters in the matching algorithm we are able to customize the number of matches. The matching algorithm calculates Euclidean distance between two features and if the distance is less than a predefined threshold, these two features are considered as a match. Finally using Random Sampling Consensus (RANSAC) algorithm ~\cite{Fischler1981} we estimated fundamental matrix. The fundamental matrix estimation process is described in section ~\ref{sec: feature tracking}.  Fig.~\ref{fig: detected_features} shows selected by RANSAC algorithm corresponding points (cyan lines). The number of inliers that satisfies the estimated fundamental matrix is shown in forth column. Last column in the table shows the ratio between the number of inliers and the average number of features detected in both images in the pair.  It can be seen from the table~\ref{tab: comparison} that the original SURF implementation constantly shows higher number of inliers in the relation to the total number of detected features.  Thus, in our further experiments we will use the original implementation rather than the OpenSURF implementation.

\begin{table}
\centering
\caption{Comparison of two SURF implementations.}
\label{tab: comparison}

\begin{tabular}{ l c c c c}
\hline\noalign{\smallskip}
Implementation &  \multicolumn{1}{p{1.8 cm}}{Detected features $I_1/I_2$ }& \multicolumn{1}{p{1.5 cm}}{Matched features} & Inliers & Ratio\\
\noalign{\smallskip}\hline\noalign{\smallskip}
OpenSurf & 1021/971 & 332 & 220 & 0.22\\
OpenSurf & 1188/1171 & 255 & 168 & 0.14\\
OpenSurf & 975/961 & 132 & 82 & 0.08\\
OpenSurf & 753/676 & 181 & 103 & 0.15\\
OpenSurf & 1021/971 & 332 & 217 & 0.22\\
OriginalSURF & 1043/982 & 355 & 260 & 0.26\\
OriginalSURF & 1307/1275 & 262 & 205 & 0.16\\
OriginalSURF & 1096/1031 & 148 & 99 & 0.09\\
OriginalSURF & 737/673 & 155 & 108 & 0.16\\
OriginalSURF & 1043/982 & 355 & 266 & 0.27\\
\noalign{\smallskip}\hline
\end{tabular}
\end{table}

Figure~\ref{fig: correspondence features} shows some examples of features detected by original and open SURF algorithms.

\section{Features extraction}

\begin{figure}
\centering
\includegraphics[width=\linewidth]{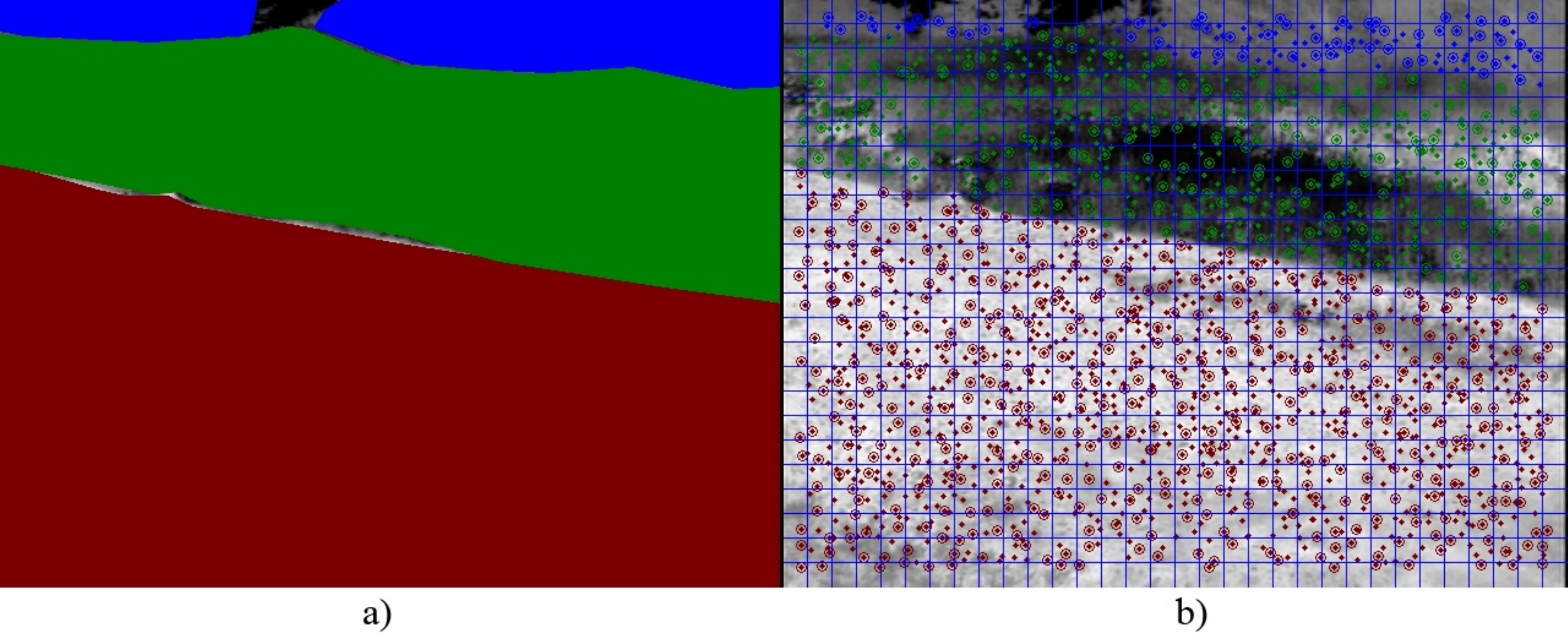}
\caption{a) Hand segmented image, b) Detected features, selected features are circled. Blue, green and dark red features represent trees, grass and road correspondingly.}
\label{fig: hand_segmented}
\end{figure}

The U-SURF consists of two steps. Firstly, using a blob detector, interest points are detected. Then, around each interest point a region descriptor is calculated. The number of extracted features is highly dependent on the image content and on the blob response threshold. The large number of points will slow down the segmentation process and more irregular regions will generate higher quantity of features leading to non-uniform feature distribution. On the other hand the small number will lead to a low spatial segmentation resolution. To maintain the number of detected features approximately constant across frames, we detect interest points at a low blob response threshold and then among them we select the strongest and uniformly distributed ones. Specifically, an image is partitioned into a grid of square cells. Strengths of points from each square are compared and the strongest feature  is selected, while others are omitted (fig.~\ref{fig: hand_segmented}(b)). Selected features are sorted into class arrays according to their labels obtained from hand segmented maps (fig. 4(a)). We considered three terrain classes: a) grass and small shrubs, b) trees and c) road including gravel and soil.  Among 3506 video frames, we selected 70 frames for training purposes and 10 for testing. Overall 45361 features were extracted from the training images.

 Lenskiy and Lee~\cite{Lenskiy2011} analyzed a number of classifiers for the purpose of SURF features classification. A classifier based on the multilayer perceptron(MLP) was found to be the most suitable in terms of classification accuracy, memory and processing time requirements. Although, the MLP is able to generalize and classify complex domains of data, some data preprocessing must be performed. The necessity comes from the fact that terrains of different classes are often mixing up making it impossible to separate one terrain region from the others even by a human. Therefore, a hand segmented region often contains fragments of a few types. To exclude features representing such fragments as well as non-informative features we omit those of them which reside far apart from the features of its own class~\cite{Lenskiy2011}. After excluding such features we are left with 41887 features, specifically 9080, 5852 and 26955 features in grass, trees and road classes correspondingly.

For the purpose of analyzing how good terrain classes are separated in the SURF features space we divided the whole training set into two subsets. The first subset consists of features surrounded by at least three/five neighbors of the same class. We call this subset a dense subset. The second subset contains all the remaining features. The number of features when at least three neighbors are of the same class is equal to 16705 (40\%), the number of remaining features is 25182 (60\%). When at least five neighbors are of the same class, there are 16494 (40\%) features in the dense subset and 25393 (60\%) features in the non-dense subsets. To check how well features from different terrain classes are separated we calculated interclass and intraclass variability as follows:

\begin{equation}
\nu (x, y) = \frac{1}{N_1 N_2} \sum^{N_1}_{i=1} \sum^{N_2}_{j=1} \sqrt{\sum^{36}_{k=1} \left( x_{i,k} - y_{j,k}\right)^2}
\end{equation}

where $x$, $y$ are feature sets, $N_1$ and $N_2$ are number of features in feature sets  $x$ and $y$ respectively.

\begin{figure}
\centering
\includegraphics[width=\linewidth]{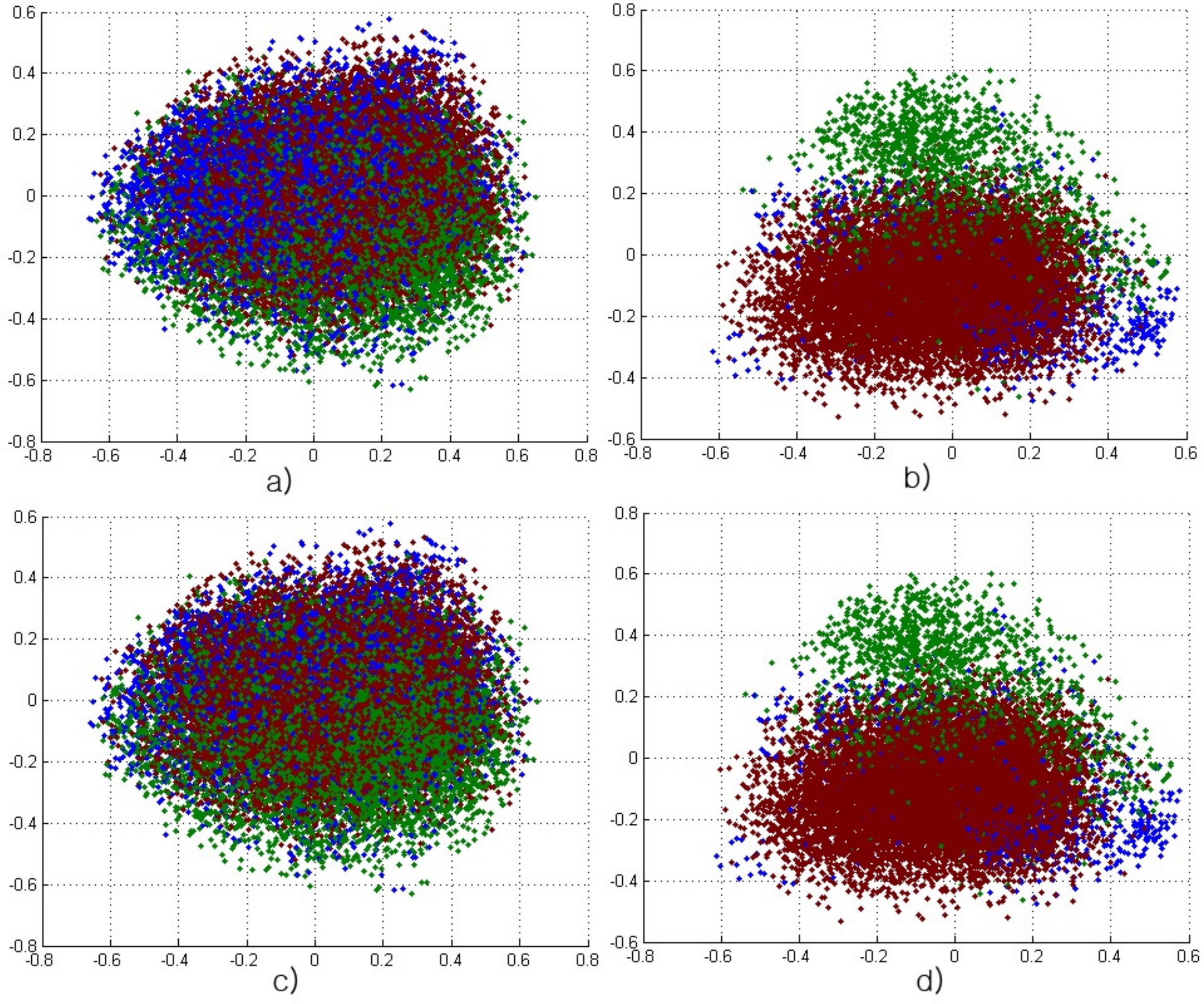}
\caption{Dimensionally reduced feature spaces. a) and c)  show non-dense feature subset, where for each feature among 5 (a) or 3 (c) its neighbors, one or more features belong to a different class. b) and d) show dense features subset, where among each feature all 5 (b) or 3 (d) neighbors belong to the same class.}
\label{fig: feature spaces}
\end{figure}

Tables ~\ref{tab: 2} show estimated variability for the whole training set, for the non-dense and dense subsets respectively. Values on the main diagonal represent interclass variabilities, and the remaining values represent variabilities between corresponding classes. As it can be seen in the cases of grass and road features, values of interclass variabilities are smaller than values for intraclass variability. However, in the case of features from the class associated with trees, the interclass variability is not smaller than the variability calculated between trees and road classes even for features from the dense-subset. This although does not necessary mean that the class is not separable at all, it could be due to non-normal feature distribution. To check if this is the case we visualize features’ distribution. We apply the principal component analysis(PCA) to each of two subsets and select among 36 components two main components. By plotting two main components of each subset we are able to visualize the features' distributions.  A solid structure is clearly recognizable, that supports the assumption that SURF features are suitable for texture classification. Secondly, as it can be seen from the plots(fig ~\ref{fig: feature spaces}, b,d), features from the tree class are split up into two clusters. Therefor a nonlinear classifier such as the MLP is needed to separate classes.

\begin{table}
  \centering
  \caption{Inter- and intraclass variance for the whole training feature set.}
  \label{tab: 2}
  \begin{tabular}{ l |c c c}
  \hline\noalign{\smallskip}
   & grass   & trees & road\\
  \noalign{\smallskip}\hline\noalign{\smallskip}
  grass & 0.79 & 0.85 & 0.77\\
  trees & 0.85 & 0.83 & 0.79\\
  road & 0.77 & 0.79 & 0.69\\
  \noalign{\smallskip}\hline
  \end{tabular}
\end{table}

\begin{table}
  \centering
  \caption{Inter- and intraclass variance for the non-dense feature subset.}
  \label{tab: 3}
  \begin{tabular}{ l |c c c}
  \hline\noalign{\smallskip}
  & grass   & trees & road\\
  \noalign{\smallskip}\hline\noalign{\smallskip}
  grass & 0.78 & 0.83 & 0.77\\
  trees & 0.83 & 0.82 & 0.79\\
  road & 0.77 & 0.79 & 0.74\\
  \noalign{\smallskip}\hline
  \end{tabular}
\end{table}

\begin{table}
  \centering
  \caption{Inter- and intraclass variations for the dense feature subset.}
  \label{tab: 4}
  \begin{tabular}{ l |c c c}
  \hline\noalign{\smallskip}
   & grass   & trees & road\\
  \noalign{\smallskip}\hline\noalign{\smallskip}
  grass & 0.81 & 0.97 & 0.83\\
  trees & 0.97 & 0.87 & 0.84\\
  road & 0.83 & 0.83 & 0.65\\
  \noalign{\smallskip}\hline
  \end{tabular}
\end{table}

\section{Texture model construction}

As it was mentioned, the number of features in each class is relatively large. To generalize and transform the data into a compact form we applied the MLP \index{multi-layer perceptron}. Following this way, features are transformed into synaptic weights of the MLP, the quantity of synaptic  weights is considerably less than the number of features multiplied by dimension of the SURF descriptor.  Therefore, significant memory reduction and reduction of processing time is achieved.

Moreover, it has been proven that a neural network with one hidden layer with sigmoid activation functions is capable of approximating any function to arbitrarily accuracy~\cite{Cybenko1989}. However, there is no precise answer on the number of neurons necessary for approximation. It was numerically shown~\cite{Chester1990} that the MLP with two hidden layers often achieves better approximation and requires fewer weights. Fewer weights mean less likeliness to get caught in local minima, while solving the optimization problem that is required in the training process. Thus, we experimented with the MLP with two hidden layers.

The training process is associated with solving a nonlinear optimization problem. Due to the fact that the error function is not convex, the training procedure is very likely to end-up in a local minimum. The error function is presented as follows:

\begin{equation}
 \varepsilon(w) = \sum_x \left( f(w, x) - class(x) \right)^2 \longrightarrow min
\end{equation}

where $f(w, x)$ is the output of the MLP, $x$ is a training set consisted of feature descriptors, and $class(x)$ desired class label obtained by hand segmentation. $class(x)$ returns a three dimensional vector, where only one component set to one, and the others are zeros.

To overcome the problem associated with the training process a number of minimization algorithms have been suggested. We experimented with "\textit{resilient propagation}" (RPROP) and the "\textit{Levenberg-Marquardt}" (LM) training algorithms~\cite{Lenskiy2011}. The former one requires less memory, but needs a large number of iterations to converge. The latter one converges in less iteration, and usually finds a better solution, however it requires a larger amount of memory. We experimented with the MLP of the following architecture: 36-40-20-3, where 36 the SURF descriptor dimension, 40 and 20 are the numbers of neurons in the first and the second hidden layers, and 3 is the number of classes. The decision function of the MLP with two hidden layer is as follows:

\begin{equation}
 f(d) = \sum_{i=0}^3 w_{m,i}^{(3)}\ \sigma \!\!\left( \sum_{j=0}^{20} w_{i,j}^{(2)}\ \sigma\!\!\left( \sum_{k=0}^{40} w_{j,k}^{(1)}\ d_k\! \right) \right)
\end{equation}

where $\sigma$ is a sigmoid activation function, $w^{(m)}$ are interlayer weight matrices. The total number of interlayer weighs is equal to $N_w = (36 + 1) \times 40 + (40 + 1) \times 20 + (20 + 1) \times 3 = 2363 $ which is considerably less than the total number of feature components $N_c = 41887 \times 36 = 1507932$.

\begin{figure}
\centering
\includegraphics[width=\linewidth]{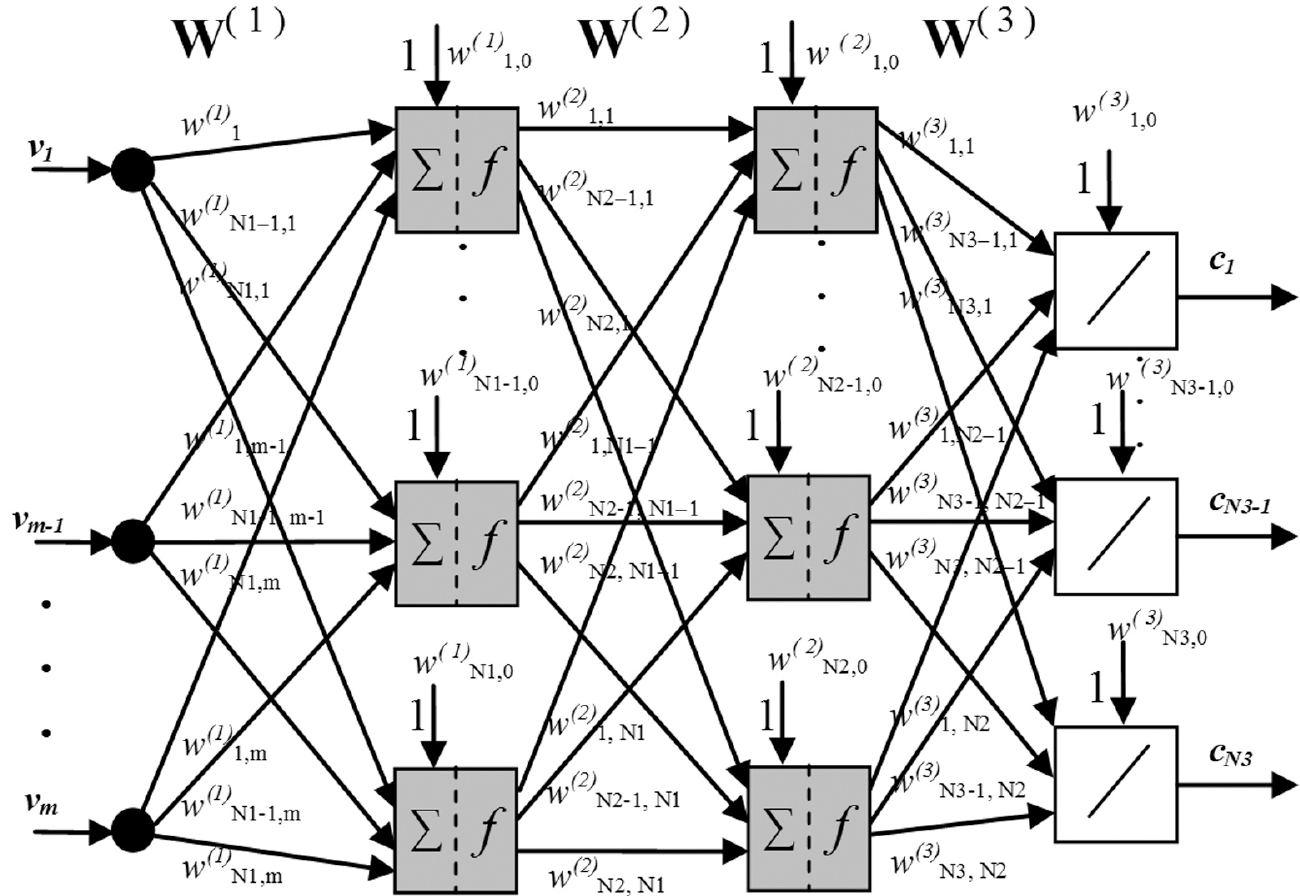}
\caption{The structure of a MLP with two hidden layers. $m=3$, $N_1 = 40$, $N_2 =20$, $N_3=3$}
\label{fig: MLP structure}
\end{figure}

\section{Segmentation procedure}

The segmentation procedure consists of two steps \index{texture segmentation algorithm}. Firstly, SURF features are extracted from the input images. The output components can be thought as class membership values. These values along with features' spatial positions $l_k=(x_k, y_k)$ are utilized by the segmentation algorithm. The algorithm is summarized in the following list of steps:
\renewcommand{\algorithmicrequire}{\textbf{Input:}}
\renewcommand{\algorithmicensure}{\textcolor{lightgray}{\textit{\ }}}
\newcommand{\OUTPUT}{\textbf{Output:}}
\newcommand{\sectionline}[1]{
%\newcommand{\WRP}{\par\qquad\(\hookrightarrow\)\enspace}
%\newline
    %\tikz \draw [lightgray, dashed] (0,0) to (0.1\linewidth,0);
    \textcolor{lightgray}{\textit{#1}}
    \newdimen{\textDimLen}%
    \setlength{\textDimLen}{\widthof{#1}}%
    %\tikz \draw [lightgray, dashed] (0,0) to (\linewidth-0.115\linewidth-\textDimLen,0);
    %\textcolor
%\newline
}

\begin{algorithm}
\caption{Segmentation algorithm}
\begin{algorithmic}[1]
\REQUIRE  $I$ - image, $d$ - feature descriptor, $l$ - feature position, $t$ - threshold.
\\
\STATE $m = \{1,2,3\}$\\
\FOR{$i=1$ to $width(I) \times height(I)$}
    \ENSURE\sectionline{Select features located within $r$ around pixel $(x_i,y_i)$}\\

    \STATE $T(x_i,y_i) = \left\{ l | \left\| (x_i,y_i)^T - l_j \right\| < r, j = 1 \dots N \right\}$ \\

    \ENSURE\sectionline{Estimate the membership value of a pixel $I(x_i,y_i)$}

    \STATE $V_{m}((x_i,y_i))=\frac{1}{\#(T((x_i,y_i)))} \sum_{k=1}^{\#(T(x_i,y_i))} d_k \cdot \frac{1}{\sigma \sqrt{2\pi}}e^{-\frac{\left\| (x_i,y_i)^T - l_k \right\| ^2}{2\sigma}}$ \\

    \ENSURE\sectionline{Assign a class index to the pixel $I(x_i,y_i)$}

    \IF{$t < \max \left( V_{m}(x_i,y_i) \right)$}
        \STATE $S(x_i,y_i) = \arg\max_{m} \left( V_{m}(x_i,y_i) \right)$ \\
    \ELSE
        \STATE $S(x_i,y_i)  = 0$ \\
    \ENDIF

\ENDFOR \\
\OUTPUT S
\end{algorithmic}
\end{algorithm}

\section{Feature tracking}
\label{sec: feature tracking}
To decrease the effect of segmentation blinking that is caused by rapid switching between different terrain types and to speed up segmentation, we match features in the current frame with features extracted in the previous frame \index{3D reconstruction}. If a corresponding point is found, the membership values calculated in the previous frame are transferred to corresponding points in the current frame. The correspondence is found by calculating Euclidean distance between feature descriptors in current and previous frames. To optimize the calculation time, the searching area in the current frame is restricted by a circle defined by a feature coordinates in the previous frame and a radius $r$. The feature with a minimum Euclidean distance in the 36-dimensional space is considered as a match if the distance is less than a predefined threshold. Based on the found matches a fundamental matrix $F$ is estimated. A fundamental matrix $F$ is the unique $3 \times 3$ rank 2 homogeneous matrix which satisfies:
\begin{equation}
m^{\prime T} F m = 0
\label{eq: fund_mat}
\end{equation}
for all corresponding points $(m,m^\prime)$. Another definition of the fundamental matrix is $l^\prime=Fm$ for the epipolar line $l^\prime$. Since $m^\prime$ belong to $l^\prime$, $m^{\prime T} l^\prime = 0 \rightarrow m^{\prime T}Fm = 0$ . We only need 8 points to solve equation (\ref{eq: fund_mat}) and find $F$. However, this solution would not be robust due to a large number of outliers. To filter them out and find fundamental matrix the RANSAC algorithm is applied~\cite{Fischler1981}. As soon as a fundamental matrix is estimated an essential matrix $E$ is calculated in the following manner:
\begin{equation}
E = W^{T}FW,
\label{eq: essential_mat}
\end{equation}
where $W$ is a matrix of intrinsic parameters. Then $E$ is decomposed ~\cite{Hartley2003} into $P= A [I|0]$, $P^\prime=A[R|t]$. $P$ and $P^\prime$ are calibrated perspective projection matrices (PPM). Using reconstructed PPMs and feature coordinates we can triangulate them and obtain 3D coordinates $M$  (fig. ~\ref{fig:  rec_scene}).

After the metric 3D reconstruction of a current camera position and orientation is provided, the prediction of the next position and orientation can be performed~\cite{Lensky2009}. Then, predicted PPM is applied to previously reconstructed 3D points $M$ to obtain a projection onto an image plane. To predict the camera’s position and orientation we use two separate trackers. For tracking the camera position we use the regular Kalman filter~\cite{Brown1997} and to predict camera orientation we applied Extended Kalman Filter, where orientation is represented using quaternion notation~\cite{Goddard1997}.

\begin{figure}
\centering
\includegraphics[width=\linewidth]{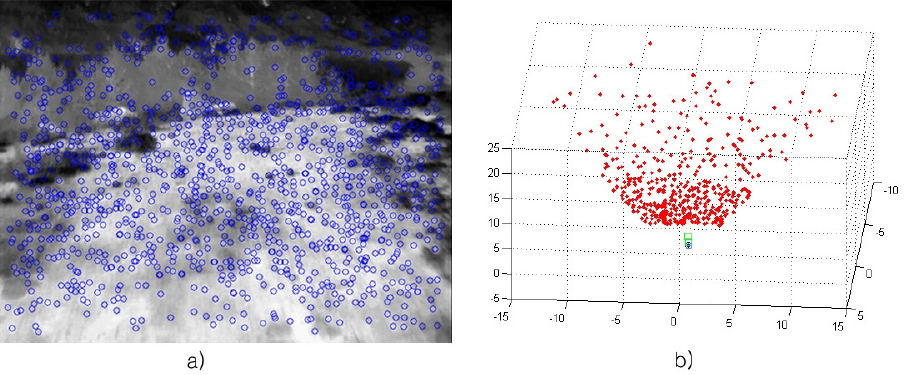}
\caption{a) Detected SURF features from a first image in the image pair, b) Reconstructed points from a pair of frames, green squares represent positions of the cameras, and the blue circle shows predicted position of the camera}
\label{fig: rec_scene}
\end{figure}

\section{Experimental results}

To test the segmentation algorithm we run experiments on two image sets: the training set and the testing set.  The training image set consists of 70 images, and the testing set contains 10 images. We experimented with he MLP trained with the \textit{RPROP} and \textit{LM} algorithms. The mean, standard deviation, minimum and maximum segmentation error rates are shown in table ~\ref{tab: 5}.

\begin{figure}
\centering
\includegraphics[width=\linewidth]{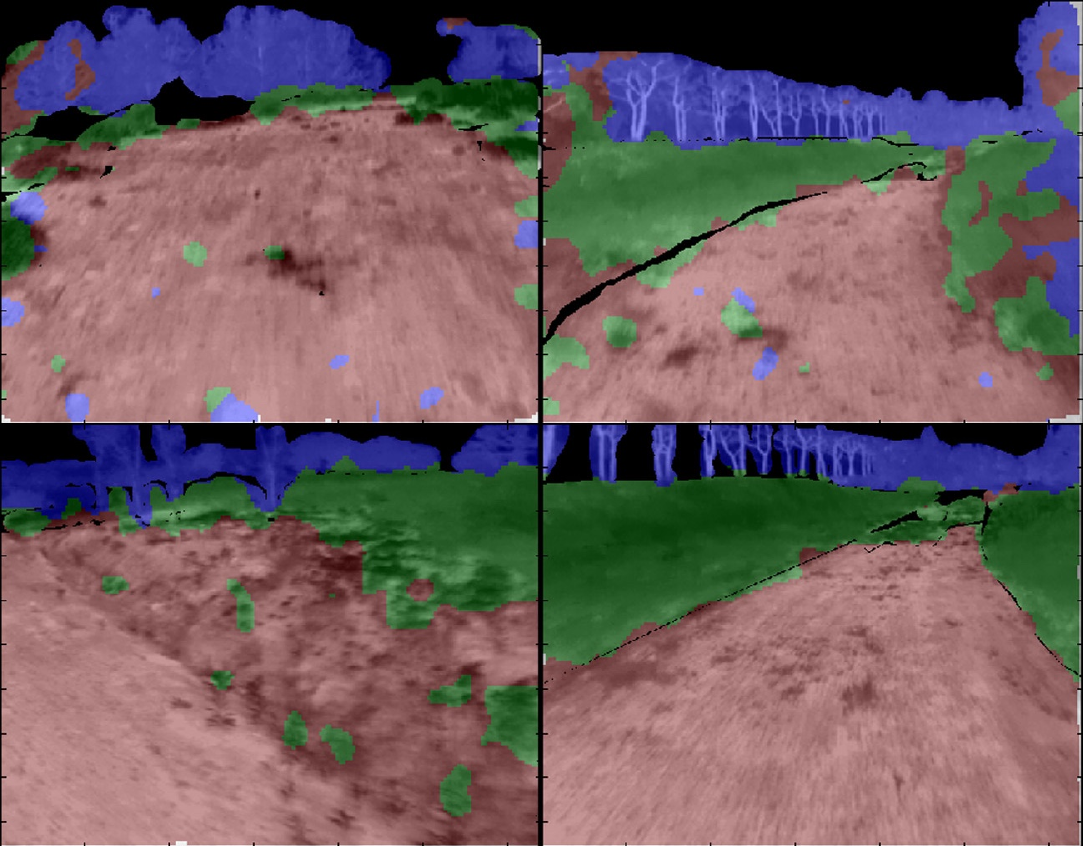}
\caption{The first and second rows show segmentation results obtained for the test and training images respectively, when nearest neighbor classifier was applied.}
\label{fig: 7}
\end{figure}

\begin{figure}
\centering
\includegraphics[width=\linewidth]{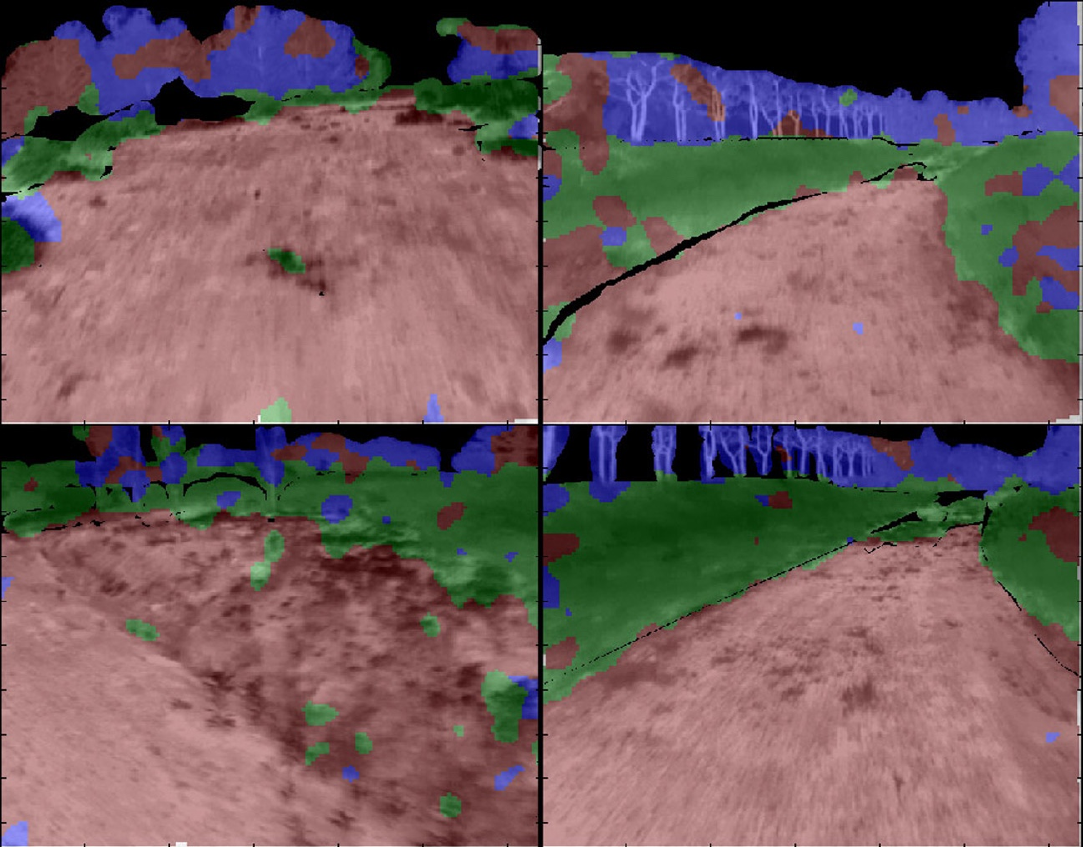}
\caption{The first and second rows show segmentation results obtained for the test and training images respectively, when MLP classifier trained by LM algorithm was applied.}
\label{fig: 8}
\end{figure}

\begin{figure}
\centering
\includegraphics[width=\linewidth]{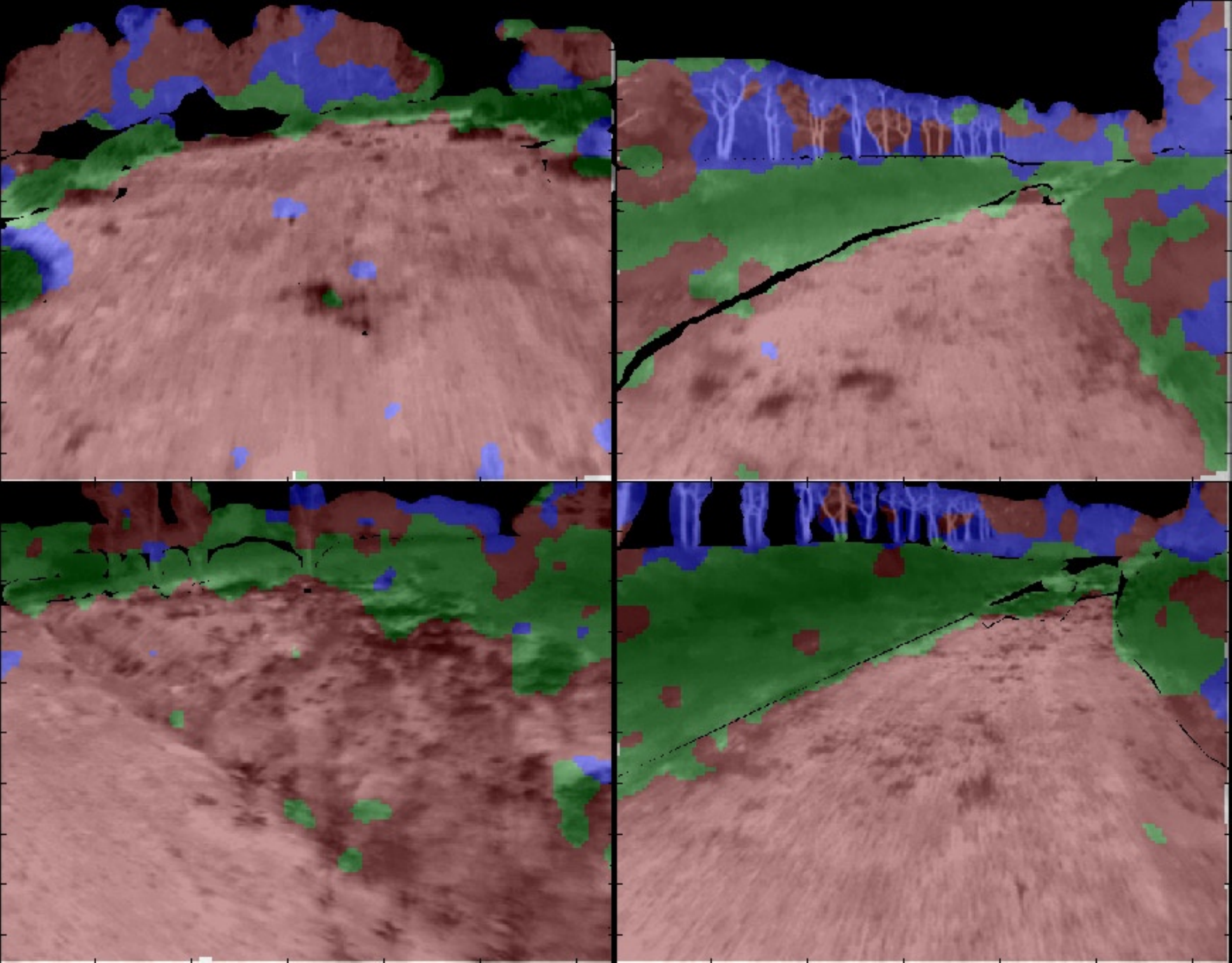}
\caption{The first and second rows show segmentation results obtained for the test and training images respectively, when MLP classifier trained by RP algorithm was applied.}
\label{fig: 9}
\end{figure}

The lowest error rate is achieved with the nearest neighbor classifier (NN). The NN classifier looks for a nearest feature and classifies a feature of interest to  the nearest feature$^\prime$s class. The distance to the nearest feature is used to calculate the strength associated with the assigned class~\cite{Lenskiy2011}. The $NN$ classifier uses during classification all training samples, as a result the segmentation error on the training set is the lowest compared to the MLP classifier.  On the other hand MLP converts the training set into a compact set of interlayer weights that considerably reduce memory requirements and decreases calculation time. The MLP trained by the $LM$ algorithm shows comparably lower segmentation error rates followed by the MLP trained by $RPROP$. In ~\cite{Lenskiy2010b} a network with 40 neurons in each hidden layer was trained with the LM algorithm. Here we reduced the number of neurons in the second hidden layer to 20. We noticed that each time the network is retrained different segmentation accuracy is obtained. Moreover the increase of number of neurons up to 60 neurons in each hidden layer (experiments not shown) did not lead to better quality of segmentation. On reason, is that the learning algorithms gets stuck in a local minimum as more variable/weights makes the optimization problem more complex. Another reason is poorer generalization of a network with larger number of weights. In this case the network memorizes noisy data.
The segmentation results obtained with the NN and MLP trained by the LM and RPROP algorithms are shown in figures ~\ref{fig: 7},~\ref{fig: 8} and ~\ref{fig: 9}.

\begin{table}
  \centering
  \caption{Error rates using different classifiers and data sets}
  \label{tab: 5}
  \begin{tabular}{ l l l l l}
  \hline\noalign{\smallskip}
  Classifier & \multicolumn{1}{c}{mean} & \multicolumn{1}{c}{std} & \multicolumn{1}{c}{min} & \multicolumn{1}{c}{max}\\
  \noalign{\smallskip}\hline\noalign{\smallskip}
  NN (training set) & 5.03 & 1.66 & 2.89 & 10.7\\
  MLP 40-20 LM(training set) & 15.25 & 5.26 & 7.27 & 29.31\\
  MLP 40-20 RPROP(training set) & 18.90 & 6.42 & 8.95 & 37.30 \\
  NN (test set) & 16.67 & 9.17 & 9.86 & 32.1\\
  MLP 40-20 LM (test set) & 19.1 & 10.39 & 12.5 & 37.52\\
  MLP 40-20 RPROP(test set) & 21.51 & 9.72 & 13.14 & 38.29\\
  \noalign{\smallskip}\hline
  \end{tabular}
\end{table}

The lowest error rate is achieved with the nearest neighbor classifier (NN). The MLP trained by LM algorithm shows comparably low segmentation error rate followed by the MLP trained by RPROP. In~\cite{Lenskiy2010b} a network with 40 neurons in each hidden layer was employed. Here we reduced the number of neurons in the second hidden layer to 20. We noticed that each time the network is retrained different segmentation accuracy is obtained.

\section{Conclusions}

Besides comparing the original and the open SURF implementations, we found that the ratio of the number of extracted features to the number of correctly detected corresponding points is higher for upright version of the original implementation than for the OpenSURF implementation. We demonstrated that 36 dimensional SURF features indeed carries meaningful information applicable for terrain segmentation. We applied segmentation algorithm that extracts 36 dimensional SURF features and estimates class membership values for each pixel.  Three terrain classes: grass, bushes and trees have been considered. The proposed segmentation approach effectively supplements those vision systems that use salient features for 3D reconstruction and object recognition. Among classification algorithms we tested the MLP trained by the $LM$ and $RPROP$ training algorithms, and the nearest neighbour algorithm. The segmentation system based on the network estimator shows slightly higher segmentation error rates compared to the estimator based on nearest neighbor algorithm. Nevertheless, the number of interconnecting weights is significantly lower than the number of training samples, which results in considerable memory and computational savings.

\end{document}